%% file: main.tex
\documentclass[runningheads]{llncs}

\usepackage{eccv}

\usepackage{eccvabbrv}

\usepackage{amssymb}
\usepackage{xcolor}
\usepackage{caption}
\usepackage{booktabs}
\usepackage{graphicx}
\usepackage{colortbl}
\usepackage{multirow}
\usepackage{subcaption}
\usepackage{pifont}
\usepackage{bm}

\usepackage[accsupp]{axessibility}  

\usepackage{hyperref}

\usepackage{orcidlink}

\begin{document}

\title{FLUX3D: High-Fidelity 3D Gaussian Generation with Diffusion-Aligned Sparse Representation} 

\titlerunning{FLUX3D}

\author{Haorui Ji\inst{1} \and
Weizhe Liu\inst{2}\and
Hongdong Li\inst{1}\and
Hengkai Guo\inst{2}}

\authorrunning{H.~Ji et al.}

\institute{The Australian National University \and
ByteDance}

\maketitle

\begin{center}
    \centering
    \captionsetup{type=figure}
    \includegraphics[width=\textwidth]{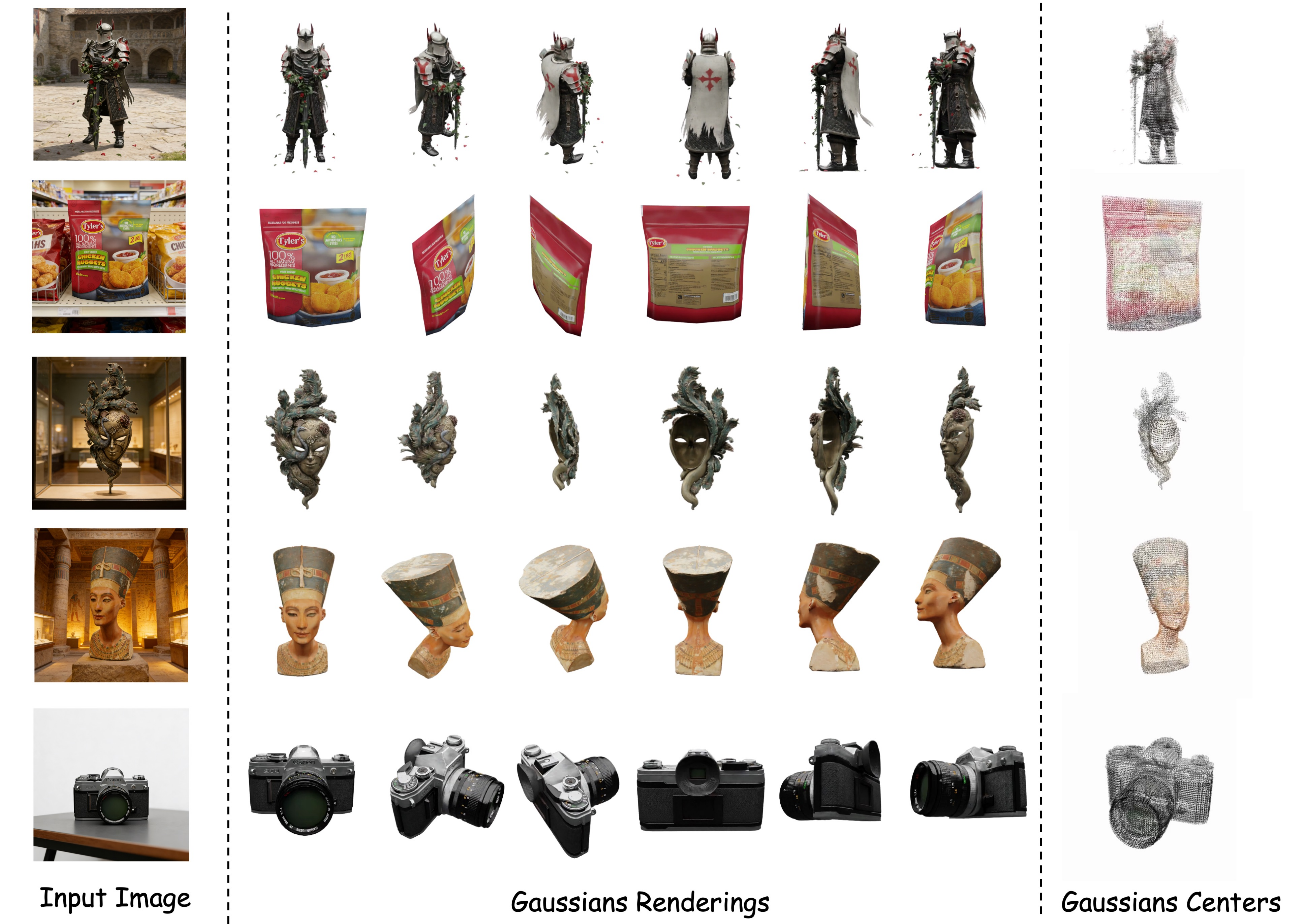}
    \captionof{figure}{We present \textbf{FLUX3D}, a scalable image-to-3DGS generation framework based on diffusion-aligned sparse voxel representation that achieves high appearance fidelity. The figure shows, from left to right, the input 2D image, 360-degree renderings of the generated 3D Gaussians, and a visualization of the centers of all Gaussians that compose the 3D assets.}
    \label{fig:teaser}
\end{center}%

\input{section/0_abstract}

\input{section/1_intro}
\input{section/2_related_work}
\input{section/3_method}
\input{section/4_exp}

\input{section/5_conclusion}

%
%
\bibliographystyle{splncs04}
\bibliography{main}

\title{FLUX3D: High-Fidelity 3D Gaussian Generation with Diffusion-Aligned Sparse Representation \\
(\textit{Supplementary Materials})}
\author{}
\institute{}
\titlerunning{FLUX3D}
\authorrunning{H.~Ji et al.}
\maketitle
\input{section/X_suppl}

\end{document}

%% file: section/0_abstract.tex
\begin{abstract}
Sparse voxel representation has emerged as a scalable foundation for image-to-3D Gaussian Splatting (3DGS) generation, yet current methods struggle to preserve high-frequency visual details of input images due to two structural bottlenecks. First, they adopt discriminative 2D features optimized for semantic abstraction to construct sparse voxel latents, which suppress reconstructive cues and induce a representation bottleneck. Second, in the generation stage, standard diffusion transformers lack effective mechanisms to align dense 2D image tokens with sparse 3D voxel latents, resulting in a cross-modal correspondence bottleneck. To address these issues, we propose FLUX3D, a scalable image-to-3DGS framework that boosts both representation learning and cross-modal alignment during generation. We first revisit 2D feature selection for sparse-voxel-based 3D representation learning, propose Diffusion-Aligned Structured Latents (DA-SLAT) and couple it with a decoder-only architecture to improve 3DGS reconstruction fidelity. We also design a sparse-structure-aware diffusion framework, which integrates the Sparse-structure Multimodal Diffusion Transformer (SMDiT) and Modal-Aware Rotary Positional Embedding (MARoPE) to achieve geometry-agnostic 2D-3D alignment. Extensive benchmark experiments demonstrate that FLUX3D yields substantial improvements in appearance fidelity and significantly outperforms all state-of-the-art (SOTA) methods in generating high-quality 3DGS assets.

\keywords{Image-to-3D Generation \and Gaussian Splatting \and High-Fidelity Appearance Modeling}

\end{abstract}

%% file: section/1_intro.tex
\section{Introduction}
\label{sec:intro}
Image-to-3D generation has emerged as a central problem in 3D content creation, aiming to synthesize plausible 3D assets from a single image of arbitrary objects. Typical pipelines follow the "representation learning + latent diffusion" paradigm, and recent progress in expressive 3D representations~\cite{zhang20233dshape2vecset,xiang2025structured} and large-scale generative foundation models~\cite{zhang2024clay,li2025triposg,li2025craftsman3d,tang2024lgm} has greatly advanced this field.

A pivotal advancement in this direction is the unification of geometry and texture generation via sparse voxel representations~\cite{ren2024xcube,xiang2025structured}. This paradigm encodes 3D objects into a sparse voxel grid, where a 3D occupancy mask defines valid spatial regions and each active voxel is associated with a latent feature vector for geometric and texture information. Compared with dense 3D representations, it reduces computational and memory overhead while supporting both global and local structure modeling, making it widely adopted by a series of SOTA 3D generation frameworks~\cite{wu2025direct3d,li2025sparc3d,chen2025ultra3d}.

Despite their strengths in capturing geometric structures and generating high-resolution shapes, sparse-voxel-based methods struggle to preserve fine-grained appearance details from input images, often leading to blurred textures and loss of high-frequency visual patterns. We argue that these artifacts stem not only from the capacity limitation of diffusion models, but also from deeper mismatches in how 2D information is transferred and aligned within the 3D latent space. On one hand, in the representation learning stage, sparse voxel latents are typically constructed with discriminative 2D features (e.g., DINOv2~\cite{oquab2023dinov2}), which are optimized for semantic abstraction rather than reconstructive fidelity, attenuating high-frequency appearance cues and creating a representation bottleneck. On the other hand, during the generation stage, dense 2D image tokens and sparse 3D voxel latents are not effectively aligned. The naive application of attention mechanisms and positional encoding designs in standard diffusion transformers does not explicitly account for the spatial topology and structural sparsity inherent in the sparse-voxel latent space. This often leads to suboptimal cross-modal interaction and creates a bottleneck that impairs visual consistency between 2D inputs and generated 3D assets.

To address these two bottlenecks in a unified framework, we propose FLUX3D, an image-to-3DGS generation framework dedicated to high-fidelity appearance reconstruction. We adopt 3DGS as the final 3D representation for its advantages in real-time photorealistic rendering. For the representation bottleneck, we revisit 2D feature selection for sparse-voxel-based 3D representation learning and hypothesize that pretrained generative diffusion features~\cite{flux2024}——optimized for image reconstruction/synthesis and rich in high-frequency appearance details——are a more suitable latent foundation for appearance-preserving 3D modeling. Based on this, we propose Diffusion-Aligned Structured Latents (DA-SLAT) coupled with a decoder-only architecture. This design leverages pretrained FLUX diffusion features as structured latents and directly maps them to 3DGS outputs, eliminating unnecessary compression and enabling faithful appearance reconstruction. For the cross-modal alignment bottleneck, inspired by SOTA 2D diffusion methods~\cite{esser2024scaling,flux2024,batifol2025flux}, we introduce a sparse-structure-aware diffusion framework consisting of two key components: (1) Sparse-structure Multimodal Diffusion Transformer (SMDiT), which explicitly accounts for sparse voxel topology and incorporates both modality-specific and joint attention processing to facilitate cross-modal information exchange; (2) Modality-Aware Rotary Positional Embedding (MARoPE), which embeds 2D image tokens onto a 3D virtual plane while preserving native voxel coordinates, enabling implicit 2D-3D correspondence learning without explicit geometric alignment.

To summarize, the key contributions of our work are as follows:
\begin{itemize}
    \item We analyze the impact of 2D feature selection in sparse-voxel-based 3D representation learning, propose DA-SLAT, and couple it with a decoder-only architecture to directly bridge diffusion features and 3DGS outputs, effectively boosting reconstruction fidelity.
    \item We design SMDiT and MARoPE as complementary components for the sparse-structure-aware diffusion framework, which collectively enhance cross-modal alignment between sparse 3D voxel latents and dense 2D image tokens.
    \item Extensive quantitative and qualitative experiments on standard image-to-3D benchmarks show that our approach generates 3D assets with high-fidelity, image-consistent appearances, and outperforms SOTA methods across key evaluation metrics.
\end{itemize}

%% file: section/2_related_work.tex
\section{Related Works}
\label{sec:related works}

\subsection{Image-to-3D Generation}
\noindent\textbf{3D Generation from 2D Guidance.}
The breakthroughs of 2D generative models have catalyzed extensive research on leveraging their priors to facilitate 3D generative tasks. DreamFusion~\cite{poole2022dreamfusion} pioneers this direction by introducing Score-Distillation-Sampling (SDS), a key technique that aligns 3D representation rendering with the score distribution of a 2D diffusion model. Subsequent works extend the SDS-based pipeline to specialize in image-to-3D generation~\cite{tang2023make,liu2023zero,gao2024cat3d,liu2023one}. Despite these advancements, 2D-guided approaches remain constrained relative to native 3D models as they require computationally expensive optimization and lack explicit 3D geometric awareness.

\noindent\textbf{Deterministic 3D Reconstruction Models.}
A parallel line of research focuses on native 3D models for image-to-3D tasks, with one prominent branch being large-scale \textit{deterministic reconstruction models}. These methods aim to reconstruct 3D assets in an end-to-end feed-forward manner, eliminating the need for lengthy iterative optimization. 
LRM~\cite{hong2023lrm} lays the foundation for this direction with the first transformer-based architecture, unleashing the scalability of large-scale datasets and model parameters. Extending beyond single-view inputs, many successors integrate multi-view image cues~\cite{zhang2024geolrm,xu2024grm,tang2024lgm}.

However, deterministic reconstruction models are inherently unable to model the uncertainty in image-to-3D task—they map 2D inputs to a fixed 3D output, rather than a distribution of plausible interpretations. This limitation often results in unsatisfactory predictions, such as blurry textures, distorted geometry, or inconsistent details.

\noindent\textbf{Stochastic 3D Generation Models.}
To address the ambiguities in image-to-3D task, another branch of native 3D models adopts the \textit{stochastic generation paradigm} based on diffusions to model the distribution of plausible 3D assets. 

Stochastic 3D generation models can be categorized by their choice of latent representations, with two mainstream families: \textit{vecset-based} and \textit{voxel-based}. Vecset-based methods~\cite{zhang20233dshape2vecset,zhang2024clay,li2025triposg,li2025craftsman3d} encode 3D shapes using sets of latent vectors (implicit representations) that can be decoded into neural signed distance functions or occupancy fields, demonstrating their ability to capture complex 3D shape variations. 

Voxel-based methods leverage explicit spatial structures in the latent space to balance expressiveness and efficiency. Recent advances~\cite{ren2024xcube,xiang2025structured,li2025sparc3d,chen2025ultra3d,wu2025direct3d} encode 3D objects by distributing latent features across a sparsely populated voxel grid. This design facilitates efficient feature aggregation within local 3D neighborhoods, reducing computational overhead compared to dense voxel representations while retaining geometric awareness. Despite these strengths, existing sparse voxel-based models often struggle to preserve fine-grained textural details from the input image, due to challenges in aligning sparse 3D latents with dense 2D image pixels—a gap that motivates our research.

\subsection{Denoising Diffusion and Flow Matching}
Denoising diffusion probabilistic models and flow matching~\cite{ho2020denoising,song2020score,lipman2022flow,liu2022flow} have emerged as transformative paradigms for generative modeling. By learning to reverse a gradual degradation process or modeling continuous vector fields between distributions, these approaches have achieved SOTA performance in high-fidelity content creation, flexible conditional generation, and scalable inference across various domains, such as image and video synthesis~\cite{wu2025qwen,seedream2509seedream,gao2025seedance}, controlled editing~\cite{wang2025unicombine,tan2025ominicontrol}, dense predictions~\cite{ke2024repurposing,gui2024depthfm,ji2024dpbridge}, and 3D applications~\cite{xiang2025structured,zhang2024clay,ji2024jade,ji20243d}. In this paper, we apply rectified flow to 3D generation and adapt its model architecture for sparse voxel data structure to improve cross-modal consistency.
\vspace{-1em}

%% file: section/3_method.tex
\begin{figure*}[t]
    \centering
    \includegraphics[width=\textwidth]{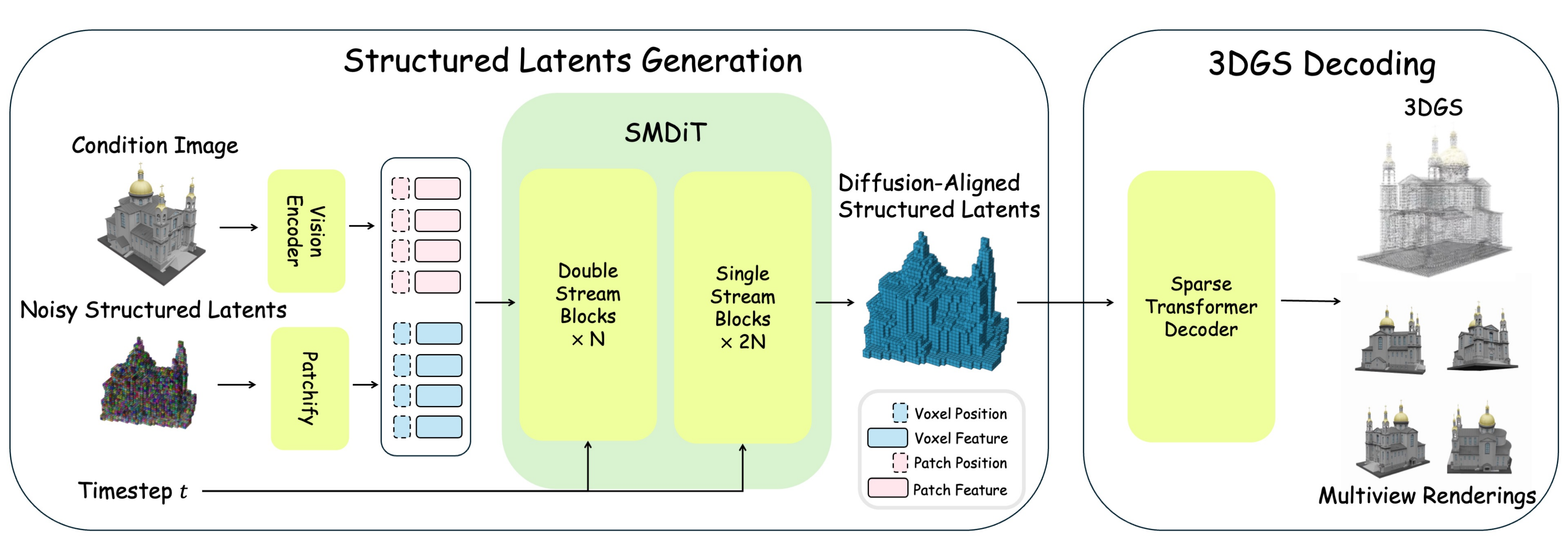}
    \caption{Overview of FLUX3D for 3D Gaussian generation. Given the condition image and sparse structure layout, we use SMDiT module to generate diffusion-aligned structured latents. The obtained latents can then be transformed into target 3D Gaussians by a sparse transformer-based decoder.}
    \label{fig:overall framework}
\end{figure*}

\section{Preliminaries}
\label{sec:preliminary}
\noindent\textbf{Sparse Voxel-based Representation.}
A 3D asset is represented as a set of feature-coordinate pairs:
\begin{equation}
    \{(\bm{f_i}, \bm{p_i})\}^L_{i=1}, 
    \quad \bm{p_i} \in \{0, . . . , N-1\}^3, 
    \quad \bm{f_i} \in \mathbb{R}^C
\end{equation}
where $\bm{p_i}$ denotes the spatial index of an active voxel within a cubic grid of resolution $N$, and $\bm{f_i}$ incorporates its associated feature. Since the number of active voxels $L$ is typically much smaller than the full grid size ($L \ll N^3$), this sparse representation enables efficient high-resolution modeling while preserving spatial locality of both structural and appearance information. The resulting feature volume can be flexibly decoded into downstream 3D representations via task-specific heads.

Sparse-voxel-based 3D generation typically follows a two-stage pipeline. Stage one generates the spatial arrangement of active voxels $\{\bm{p_i}\}^L_{i=1}$, while in stage two, corresponding features $\{\bm{f_i}\}^L_{i=1}$ that encode shape and appearance information are derived. To enhance computational efficiency, both stages of generation are conducted in the latent space. In this paper, we assume that the sparse voxel layout is given, and our main focus is to produce better features to improve appearance fidelity.

\noindent\textbf{3D Gaussian Splatting.} 3DGS~\cite{kerbl20233d} is a real-time radiance field representation that models 3D assets via a collection of discrete 3D Gaussian primitives, each parameterized by $\bm{G} = \{\boldsymbol{\mu}, \bm{s}, \bm{r}, \alpha, \bm{c}\}$. Here, $\boldsymbol{\mu} \in \mathbb{R}^3$ denotes the center position; $\bm{s} \in \mathbb{R}^3$ and $\bm{r} \in \mathbb{R}^3$ are the scale and rotation that describe the ellipsoid shape; $\alpha \in \mathbb{R}$ stores the opacity and $\bm{c} \in \mathbb{R} ^d$ represents color information via spherical harmonics. During rendering, 3D Gaussians are first projected onto the 2D image plane, followed by front-to-back alpha compositing to blend the color and opacity of overlapping Gaussians, producing final pixel values. In our setting, sparse voxel features are decoded into 3DGS parameters as the final asset representation.

\section{Method}
\label{sec:method}
In this section, we detail the framework design of FLUX3D, which aims to generate high-fidelity 3DGS from a single input image under the sparse voxel representation. As discussed earlier, appearance degradation in existing sparse-voxel-based pipelines arises from two structural bottlenecks: (1) the use of discriminative 2D features that suppress reconstructive cues during structured latents construction, and (2) insufficient alignment between dense 2D image tokens and sparse 3D voxel latents during generation. As illustrated in~\cref{fig:overall framework}, our framework addresses these two issues through improved representation learning and sparse-structure-aware multimodal diffusion modeling.

\begin{figure}[t]
    \centering
    \setlength{\abovecaptionskip}{0.3em}
    \includegraphics[width=\linewidth]{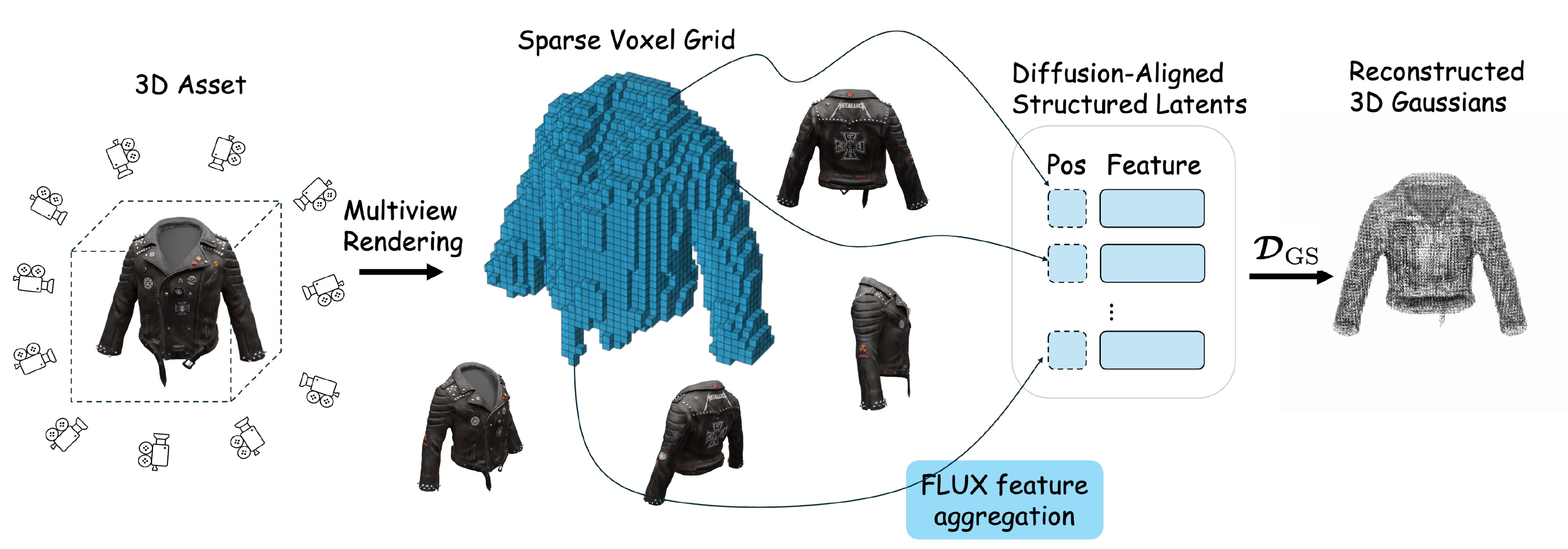}
    \caption{An illustration of our representation learning pipeline. We extend TRELLIS-SLAT by two modifications: we use FLUX features (as opposed to DINOv2 features) aggregated from multiview renderings to construct diffusion-aligned structured latent space, and use a decoder-only architecture (as opposed to a standard encoder-decoder one) to directly map the structured latent space to 3DGS output.}
    \label{fig:da-slat}
\vspace{-1em}
\end{figure}

\subsection{Diffusion-Aligned Structured Latents}
\label{sec:vae}
We adapt Trellis-SLAT~\cite{xiang2025structured} to our representation learning pipeline. As shown in~\ref{fig:da-slat}, we render multiview images from spherically sampled cameras and voxelize the input 3D asset into a feature volume $\{(\bm{f_i}, \bm{p_i})\}^L_{i=1}$, where the feature of each active voxel is derived by aggregating pre-extracted features from the multiview images. In its original formulation, Trellis-SLAT utilizes DINOv2 features~\cite{oquab2023dinov2} to construct the feature volume $\{(\bm{f_{i,dino}}, \bm{p_i})\}^L_{i=1}$ and employs a sparse transformer encoder-decoder architecture for structured latent learning and 3DGS reconstruction:
\begin{equation}
\begin{aligned}
    \bm{\mathcal{E}_{\text{trellis}}}&: \{(\bm{f_{i,\textbf{dino}}}, \bm{p_i})\}^L_{i=1} \to \{(\bm{z_i}, \bm{p_i})\}^L_{i=1} \\
    \bm{\mathcal{D}_{\text{GS,trellis}}}&: \{(\bm{z_i}, \bm{p_i})\}^L_{i=1} \to \{\{\bm{G}_i^k\}_{k=1}^K\}_{i=1}^L
\end{aligned}
\end{equation}
where each voxel is decoded into $K$ Gaussians. To alleviate the representation bottleneck and improve appearance reconstruction fidelity, we extended the original design with two modifications.

\noindent\textbf{FLUX-based Structured Latents.} 
We first replace DINOv2 with FLUX features~\cite{flux2024} to construct the voxelized feature volume$\{(\bm{f_{i,flux}}, \bm{p_i})\}^L_{i=1}$. Although discriminative features such as DINOv2 are semantically expressive, they are optimized for semantic abstraction and invariance, which sacrifices the high-frequency appearance information critical for faithful 3D reconstruction. Moreover, aggressive dimensional compression during latent encoding ($\bm{f_{i,\textbf{dino}}} \in \mathbb{R}^{1024} \to \bm{z_i} \in \mathbb{R}^{8}$) introduces additional information loss, which further degrades the appearance fidelity of the decoded 3D assets. In contrast, generative diffusion features like FLUX are inherently optimized for image reconstruction and synthesis, and therefore retain richer appearance statistics.

\noindent\textbf{Sparse Transformer Decoder-only Architecture.}
Next, instead of adopting a conventional encoder–decoder VAE architecture, we directly learn a sparse transformer decoder that maps the FLUX-based structured latent space to 3DGS parameters:
\begin{equation}
    \bm{\mathcal{D}_{\text{GS, flux3d}}}: 
    \{(\bm{f_{i,\textbf{flux}}}, \bm{p_i})\}_{i=1}^L 
    \rightarrow 
    \{\{\bm{G}_i^k\}_{k=1}^K\}_{i=1}^L.
\end{equation}
In this way, the pre-trained FLUX diffusion features are directly used as 3D structured latents without being re-encoded into a separate compact representation. Since diffusion features are inherently information-dense and aligned with the dynamics of generative modeling, our decoder-only design obviates the need for an additional encoder to enforce latent compression and distribution alignment. As a result, it eliminates the information loss in the traditional encoder-decoder architecture and can maintain stronger appearance fidelity while remaining compatible with the downstream diffusion-based 3D generation stage.

\subsection{Sparse-Structure-Aware Diffusion}
\label{sec:diffusion}
With a diffusion-aligned structured latent in place and demonstrating good 3DGS reconstruction quality, the generation stage focuses on modeling its conditional distribution given a reference image $\bm{I}$ and sparse voxel layout $\{\bm{p_i}\}$:
\begin{equation}
    p(\{\bm{f_{i,flux}}\}_{i=1}^L \mid \{\bm{p_i}\}_{i=1}^L, \bm{I}).
\end{equation}
A key challenge lies in aligning dense 2D image tokens with sparse 3D voxel latents. Standard diffusion transformers treat all tokens homogeneously and fail to account for the sparsity and volumetric topology of 3D voxels, leading to weak cross-modal correspondence and inconsistent appearance modeling. To alleviate this cross-modal alignment bottleneck, we introduce sparse-structure-aware multimodal diffusion, which comprises two two complementary structural improvements that work synergistically.

\begin{figure}[t]
    \centering
    \begin{minipage}[t]{0.47\textwidth}
        \centering
        \setlength{\abovecaptionskip}{0.6em}
        \includegraphics[width=\linewidth]{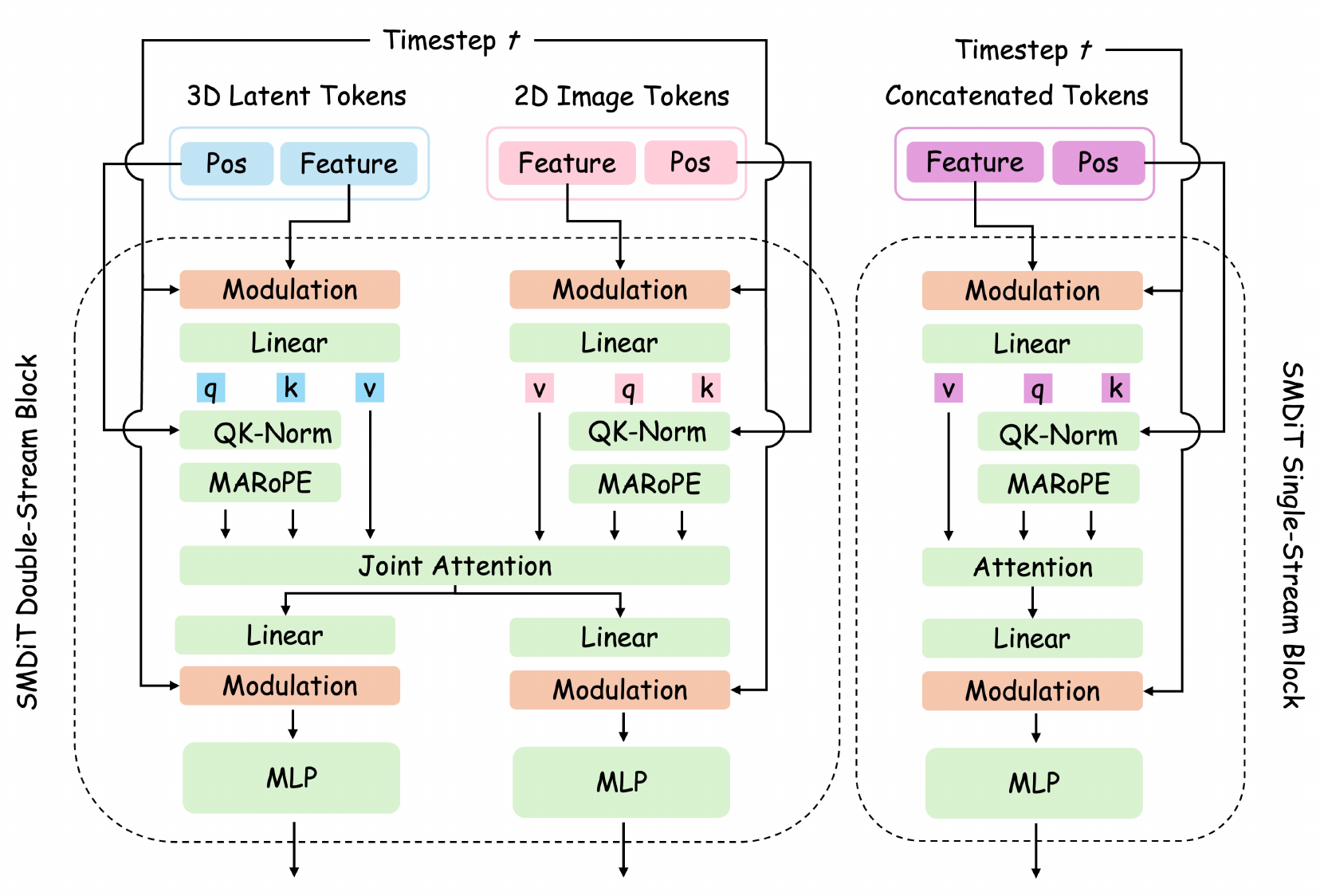}
        \caption{An illustration of the SMDiT network architecture. SMDiT employs a transformer architecture comprising both double-stream and single-stream blocks. It incorporates the characteristics of sparse-structured latent space and facilitates interactions between modalities.}
        \label{fig:smdit}
    \end{minipage}
    \hspace{0.01\textwidth}
    \begin{minipage}[t]{0.47\textwidth}
        \centering
        \setlength{\abovecaptionskip}{0.6em}
        \includegraphics[width=\linewidth]{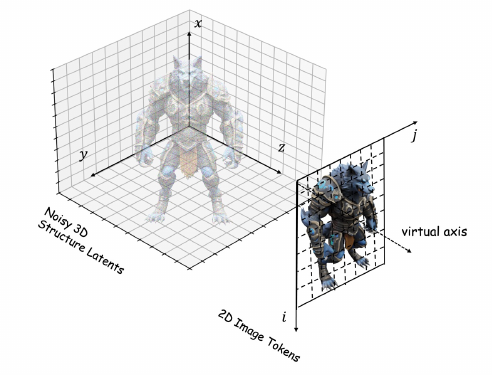}
        \caption{An illustration of the MARoPE strategy, where 2D images are mapped onto a virtual plane outside the latent volume, and raw 3D voxel coordinates within the volume are retained. This strategy enables better alignment between 3D structures and 2D images.}
        \label{fig:marope}
    \end{minipage}
\end{figure}

\noindent\textbf{Sparse-structure Multimodal Diffusion Transformer.}
Inspired by advances in image diffusion models, we propose the Sparse-structure Multimodal Diffusion Transformer (SMDiT) architecture, which extends the Multimodal Diffusion Transformer (MMDiT) design~\cite{esser2024scaling,flux2024,batifol2025flux} to effectively capture cross-modal correspondences between sparse-structured 3D latents and conditional 2D image features. As shown in~\cref{fig:smdit}, the condition image is first tokenized using a FLUX encoder. The resulting tokens are stored in a sparse voxel data structure, where each image patch corresponds to an active voxel. Next, latent tokens and condition tokens are fed into a sequence of double-stream and single-stream blocks. In double-stream blocks, the two modalities are processed in parallel with modality-specific weights to preserve their intrinsic characteristics; in single-stream blocks, all tokens are concatenated and processed with a joint attention mechanism to facilitate effective cross-modal interaction. To further improve training efficiency, we implement structured patchification: 3D latent tokens are grouped via sparse downsampling over local \(2^3\) voxel regions, and 2D image tokens are patchified with \(2^2\) partitions.

\noindent\textbf{Modal-Aware Rotary Position Embedding}. 
Previous works~\cite{su2024roformer,wu2025qwen} have highlighted the importance of positional embeddings, as they help to distinguish between modalities and learn spatial correspondences. However, existing cross-modal positional embeddings are primarily designed for text-image or image-image settings~\cite{wu2025qwen,batifol2025flux,tan2024ominicontrol} and do not directly address challenges in image-to-3D task for linking 2D image patches to 3D volumetric regions. A few works~\cite{feng2025romantex} propose 3D-aware RoPE to encode 2D-3D correspondences explicitly, but it relies on accurate camera parameters to build a canonical coordinate map. This assumption is impractical for general-purpose 3D generation, where users rarely provide calibrated cameras, leading to geometric misalignment and degraded performance. 

To address these limitations, we propose Modal-Aware Rotary Position Embedding (MARoPE), which embeds 2D and 3D tokens within a shared 3D coordinate framework without relying on explicit geometric calibration. As shown in~\cref{fig:marope}, MARoPE maps 2D image patch indices $(i, j)$ to a ``virtual plane" located outside the voxel volume at $(i,j,z_{\max}+1)$, while 3D voxel coordinates $(x,y,z)$ remain unchanged. This design conceptualizes the condition image as being attached along the third dimension of the latent volume. Rather than recovering true 3D poses, MARoPE provides a consistent relative coordinate frame so that RoPE's distance-decay attention biases each image patch toward nearby voxels, with the 2D-3D correspondence learned from data.

\subsection{Training Details}
\label{sec:training}
\noindent\textbf{Representation Module Training.}
The representation module is trained in an end-to-end manner. At each step, we randomly choose one reference view, then minimize the perceptual L1 reconstruction loss between the rendering result of the decoded output and that of the ground-truth from the same viewing angle.  Following prior work~\cite{xiang2025structured}, we incorporate geometry-aware regularization terms on Gaussian volume and opacity. The full training objective can be written as:
\begin{equation}
    \begin{aligned}
        \mathcal{L}_{\text{total}} &= \mathcal{L}_{\text{recon}} + \mathcal{L}_{\text{vol}} + \mathcal{L}_{\alpha} \\
        \mathcal{L}_{\text{recon}} &= \mathcal{L}_1 + \lambda_{\text{ssim}}(1 - \text{SSIM}) + \lambda_{\text{lpips}}\text{LPIPS} \\
        \mathcal{L}_{\text{vol}} &= \frac{1}{LK} \sum_{i=1}^L \sum_{k=1}^K \prod \boldsymbol{s}_i^k \\
        \mathcal{L}_{\alpha} &= \frac{1}{LK} \sum_{i=1}^L \sum_{k=1}^K (1 - \alpha_i^k)^2
    \end{aligned}
\end{equation}
To enhance robustness to the continuous outputs of diffusion modeling, we inject stochasticity during training by sampling from the FLUX VAE encoder's output mean $\mu$ and variance $\sigma$. Instead of using $\mu$ along to construct the structured latent volume, we incorporate the influence of $\sigma$ by adding a stochastic term $\mu + \sigma \cdot \mathcal{N}(0, \mathbf{I})$.

\noindent\textbf{Diffusion Training.}
For generative modeling, the SMDiT model is trained with conditional flow matching (CFM)~\cite{lipman2022flow}, a continuous normalizing flow framework that formulates generation as a progressive transformation of noise toward the target distribution. Given a linear interpolation forward process $z_t = (1-t)z_0 + t\epsilon$ we train the model to predict the corresponding velocity field $u_t$ of the noisy input $z_t$ with respect to the straight-line trajectory:
\begin{equation}
    \begin{aligned}
        \mathcal{L}_{\text{CFM}}(\theta) &= \|v_t(z; \theta) - u_t(z|\epsilon)\|_2^2 \\
        &= \|v_t((1 - t)z_0 + t\epsilon; \theta) - (\epsilon - z_0)\|_2^2
    \end{aligned}
\end{equation}

%% file: section/4_exp.tex
\section{Experiments}
\label{sec:exp}

In this section, we first introduce the experimental settings, including the dataset and implementation details. Then, we showcase the quantitative and qualitative performance of ours and baseline methods across both reconstruction and generation tasks. Finally, we conduct detailed ablation studies to verify the effectiveness of our design components.

\subsection{Experiment Settings}
\noindent\textbf{Datasets.} Following~\cite{xiang2025structured}, we use 3D-FUTURE~\cite{fu20213d}, ABO~\cite{collins2022abo}, HSSD~\cite{khanna2024habitat}, and Objaverse-XL~\cite{deitke2023objaverse} datasets for training and the Toys4k dataset~\cite{stojanov2021using} for quantitative and qualitative evaluation. In addition, we manually remove assets with low-quality textures and curate 360K 3D instances from the training set. Please refer to the supplementary materials for data preparation details.

\noindent\textbf{Implementation Details.}
For reconstruction experiments, we render 150 images per asset at a resolution of 512, with cameras uniformly distributed on a sphere with a radius of 2, facing the asset origin with $40^{\circ}$ FoV. For generation experiments, the camera settings remain the same, except that we render 24 images at 1024 resolution per asset as conditioning, balancing detail richness and computational efficiency. We also assume that the sparse voxel layout is given, and only generate the attached feature to evaluate the appearance modeling quality. During training, we adopt the AdamW optimizer~\cite{loshchilov2017decoupled} with a learning rate of $1 \times 10^{-4}$. During inference, we set the number of sampling steps to 50 to ensure fair comparison across all methods. All experiments are conducted on 8 NVIDIA A100 GPUs.




\begin{table}[t]
\begin{minipage}[t]{0.47\linewidth}
\centering
\footnotesize 
\setlength{\abovecaptionskip}{0.2cm}
\resizebox{0.95\linewidth}{!}{%
\renewcommand{\arraystretch}{1.2}
\begin{tabular}{cccc}
\toprule
\textbf{Method} & \textbf{SSIM}$\uparrow$ & \textbf{PSNR}$\uparrow$ & \textbf{LPIPS}$\downarrow$ \\

\midrule
GaussianAnything    &  0.9475  &  26.73  & 0.06397   \\
TRELLIS    & 0.9719  & 31.54  & 0.02964   \\

\midrule
\rowcolor[gray]{0.85}
Ours (Enc-Dec)   & \underline{0.9779} & \underline{33.80}  & \underline{0.02699} \\
\rowcolor[gray]{0.85}
Ours (Dec-Only)   & \textbf{0.9783} & \textbf{34.12}  & \textbf{0.02668} \\
\bottomrule
\end{tabular}
}
\caption{Quantitative evaluation of 3D reconstruction with other latent representations in terms of appearance fidelity on Toys4k. \textbf{Bold} and \underline{underline} denote the best and the second-best results.}
\label{tab:reconstruction_exp}
\end{minipage}
\hspace{0.01\textwidth}
\begin{minipage}[t]{0.47\linewidth}
\centering
\footnotesize
\setlength{\abovecaptionskip}{0.2cm}
\resizebox{0.85\linewidth}{!}{%
\renewcommand{\arraystretch}{1.2}
\begin{tabular}{cccc}
\toprule
\textbf{Input Feature} & \textbf{SSIM}$\uparrow$ & \textbf{PSNR}$\uparrow$ & \textbf{LPIPS}$\downarrow$ \\
\midrule
DINOv2    &  0.9719  & 31.54  & 0.02964    \\
DINOv3    &  0.9703  & \underline{32.61}  & 0.03192    \\
Raw RGB   &  0.9720  & 31.14  & 0.03794    \\
SDXL   &  \underline{0.9732}  &  32.54  &  \underline{0.02916}   \\
\rowcolor[gray]{0.85}
FLUX(Ours)   & \textbf{0.9779} & \textbf{33.80}  & \textbf{0.02699} \\
\bottomrule
\end{tabular}
}
\caption{Ablation study on input feature selection under reconstruction experiment setting (All results are reported under encoder-decoder architecture).}
\label{tab:ablation on input feature selection}
\end{minipage}
\vspace{-1em}
\end{table}

\subsection{Reconstruction Experiments}
We first evaluate the appearance reconstruction ability of our diffusion-aligned structured latents, whose performance establishes the upper bound for the framework's image-to-3D generation quality. For each asset, we randomly sample one camera perspective and calculate the SSIM, PSNR, and LPIPS metrics between the rendered images from the reconstructed 3DGS with the ground truth. We primarily compare our method against two baselines: TRELLIS~\cite{xiang2025structured} and GaussianAnything~\cite{lan2024gaussiananything}. The former serves as an example of sparse structure 3D generation without diffusion-aligned structured latent space, and the latter is an interactive latent space with a point cloud structure, also trained on large-scale data. The commonly used vecset latent representation is employed for shape reconstruction rather than appearance modeling, and thus is not the focus of this work. As shown in~\cref{tab:reconstruction_exp}, our method outperforms all baselines across all evaluated metrics. This result validates the effectiveness of incorporating 2D diffusion features in representation learning.

\begin{table*}[t]
\centering
\setlength{\abovecaptionskip}{0.6em}
\resizebox{\linewidth}{!}{
\renewcommand{\arraystretch}{1.2}
\begin{tabular}{c|ccc|cccccc}
\toprule
\textbf{Method}  & 
\textbf{SSIM}$\uparrow$ & 
\textbf{PSNR}$\uparrow$ & 
\textbf{LPIPS}$\downarrow$ &
\textbf{CLIP}$\uparrow$ &
$\textbf{FD}_{\textbf{incep}}\downarrow$ &
$\textbf{KD}_{\textbf{incep}}(\%)\downarrow$ &
$\textbf{FD}_{\textbf{dinov2}}\downarrow$ &
$\textbf{KD}_{\textbf{dinov2}}(\%)\downarrow$ \\
\midrule
LGM & 0.8940 & 19.94 & 0.1002 &  87.23  &  33.43  &  0.71  &  341.01  &  34.48 \\
GeoLRM & 0.9149 & 20.74 & 0.0961 & 86.65 &  37.66  &  2.65  &  362.23  & 36.05  \\
\midrule
GaussianAnything & 0.9055 & 20.33 & 0.1056 &  86.40  &  32.76  &  3.97  &  343.01  &  33.72   \\
DiffusionGS & 0.9355 & 22.86 & 0.08072 & 87.82 & 30.18 & 1.25 & 223.37 & 23.43   \\
TRELLIS  & 0.9558 & 25.48 & 0.04389 &  97.92    &  11.29    & \underline{0.047}   &  63.66  &  0.62  \\
\midrule
\rowcolor[gray]{0.85} 
Ours (Enc-Dec)  &  \underline{0.9615}  &  \underline{26.08}  &   0.03958  & \textbf{98.40}    & \underline{8.81}     & \textbf{0.039}   & \underline{58.43}     & \underline{0.49} \\
\rowcolor[gray]{0.85}
Ours (Dec-Only)  & \textbf{0.9653}  &  \textbf{26.26}  &   \textbf{0.03509}  & \underline{98.37}    & \textbf{8.73}     & \textbf{0.039}   & \textbf{54.92}     & \textbf{0.42} \\
\bottomrule
\end{tabular}
}
\caption{Quantitative evaluation of image-conditioned 3DGS generation on Toys4k dataset in terms of appearance fidelity of 2D renderings.}
\label{tab:generation_exp}
\vspace{-1em}
\end{table*}

\begin{figure*}[t]
    \centering
    \setlength{\abovecaptionskip}{0.2em}
    \includegraphics[width=\textwidth]{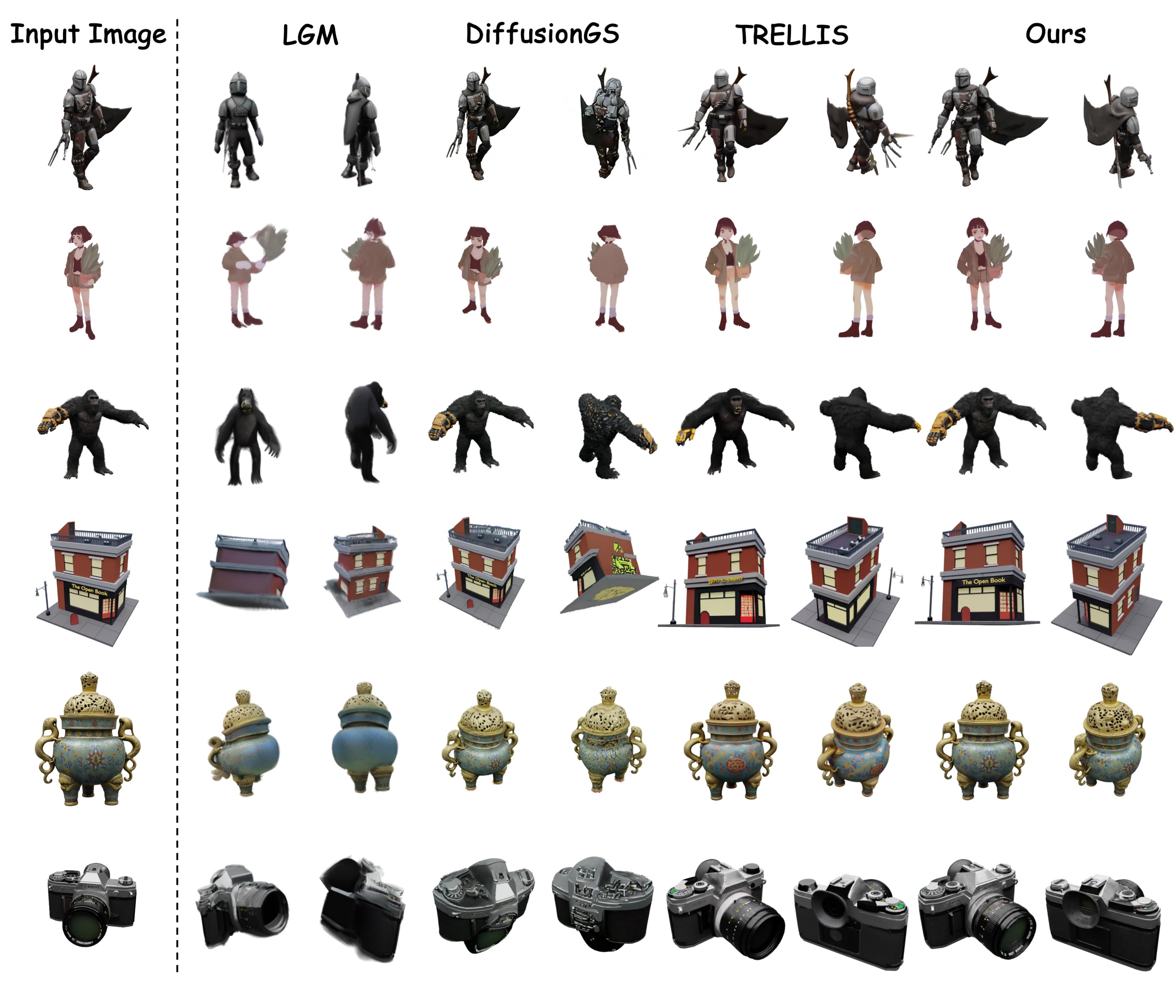}
    \caption{Visual comparisons between our method and SOTA 3DGS generation methods, including LGM, DiffusionGS, and TRELLIS. From left to right, the columns show the input 2D image, followed by the 3D reconstructions from each competing method and our approach, rendered from two viewpoints. Our method consistently produces results with high-fidelity appearance modeling across different object categories, such as characters, buildings, and artifacts. Please zoom in for detailed visualization.}
    \label{fig:qualitative}
    \vspace{-1em}
\end{figure*}
\subsection{Generation Experiments}

Regarding image-to-3DGS generation quality, we compare our proposed method with two lines of approaches: \textit{reconstruction-based}, including LGM~\cite{tang2024lgm} and GeoLRM~\cite{zhang2024geolrm}, and \textit{generation-based}, including GaussianAnything~\cite{lan2024gaussiananything}, DiffusionGS~\cite{cai2025baking}, and TRELLIS~\cite{xiang2025structured}. To assess visual consistency, we compute the SSIM, PSNR, and LPIPS between 2D renderings of the generated and ground-truth 3D assets from the input image's camera perspective. To evaluate generation quality, we also report CLIP Score, Fréchet Distance (FD), and Kernel Distance (KD), paired with distinct feature extractors. As shown in~\cref{tab:generation_exp}, our method outperforms SOTA 3DGS generation approaches across all evaluated metrics.

We further present qualitative comparisons on in-the-wild data in~\cref{fig:qualitative}. As observed in the figure, LGM exhibits significant distortions in both shape and appearance. On the other hand, DiffusionGS and TRELLIS show relatively good visual consistency from the input view, but they both suffer from inconsistent or misaligned textural details, particularly when viewed from novel perspectives. In contrast, our framework produces results that best preserve the input's color accuracy and appearance details across different object categories, such as characters, buildings, and artifacts, demonstrating its generalization capability.

\subsection{Ablation Studies}

\begin{figure*}[t]
    \centering
    \includegraphics[width=\textwidth]{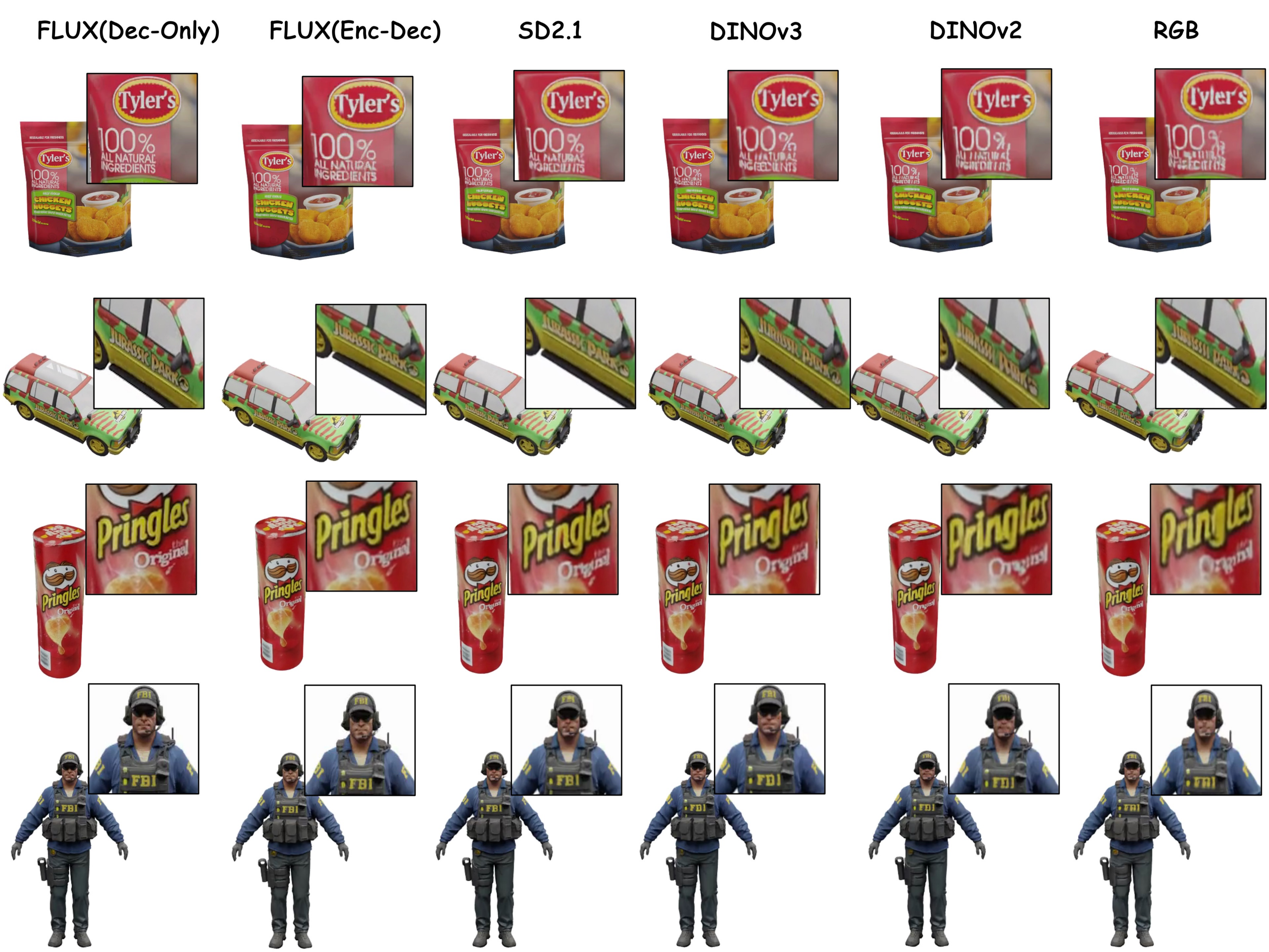}
    \caption{Visualization of the influence of input feature selection on sparse voxel representation learning. We provide zoomed-in visualizations of local regions to enable granular evaluation of detail appearance quality. We evaluate six feature extractors: FLUX (in both decoder-only and encoder-decoder configurations), Stable Diffusion 2.1 (SD2.1), DINOv3, DINOv2, and vanilla RGB pixels. Our results confirm that diffusion-aligned features, especially FLUX encodings, excel at retaining fine-grained appearance details, yielding the most visually consistent and accurate 3DGS reconstructions. This superiority is most evident in the rendering of text and logos.}
    \label{fig:ablation_rec}
    \vspace{-1em}
\end{figure*}

\noindent\textbf{Input feature selection.}
We first ablate the influence of input feature selection on representation learning. Specifically, we use different features, including DINOv2~\cite{oquab2023dinov2}, DINOv3~\cite{simeoni2025dinov3}, raw RGB pixel values, SDXL~\cite{podell2023sdxl}, and FLUX to construct the sparse structure feature volume, and evaluate their 3DGS reconstruction performance. For fair comparison, all experiments are implemented using the encoder-decoder architecture. As reported in~\cref{tab:ablation on input feature selection}, SDXL and FLUX features deliver better performance, which aligns with our observation that their reconstructive characteristics indeed help representation learning. Also shown in ~\cref{fig:ablation_rec}, diffusion-aligned features indeed improve fine-grained appearance modeling, especially in details such as text and logo rendering.

\noindent\textbf{Encoder-Decoder VS Decoder-Only.}
Next, we examine the efficacy of our decoder-only architecture compared to the encoder-decoder paradigm. As shown in the last two rows in~\cref{tab:reconstruction_exp} and ~\cref{tab:generation_exp} and the first two columns of visualization in ~\cref{fig:ablation_rec}, the decoder-only architecture outperforms the encoder-decoder one in both reconstruction and generation performance, indicating its effectiveness when paired with diffusion-aligned structured latent space.

\begin{table*}[t]
\centering
\setlength{\abovecaptionskip}{0.6em}
\resizebox{\linewidth}{!}{
\renewcommand{\arraystretch}{1.2}
\begin{tabular}{cccccccc}
\toprule
\multicolumn{3}{c}{\textbf{Configurations}} & \multirow{2}{*}{\textbf{CLIP}$\uparrow$} & \multirow{2}{*}{\textbf{FD}$_{\textbf{incep}}$$\downarrow$} & \multirow{2}{*}{\textbf{KD}$_{\textbf{incep}}$(\%)$\downarrow$} & \multirow{2}{*}{\textbf{FD}$_{\textbf{dinov2}}$$\downarrow$} & \multirow{2}{*}{\textbf{KD}$_{\textbf{dinov2}}$(\%)$\downarrow$}  \\
DA-SLAT & SMDiT & MARoPE & & & & & \\  
\midrule
\ding{55} & \ding{55} & \ding{55}  & 97.92    & 11.29    & 0.047   & 63.66  & 0.62  \\
\ding{51} & \ding{55} & \ding{55}  & 98.09    & 10.12    & \underline{0.042}  & 61.26  & 0.58  \\
\ding{51} & \ding{51} & \ding{55}  & \underline{98.15}  & \underline{9.62}  & 0.043  & \underline{60.07}  & \underline{0.55}  \\
\rowcolor[gray]{0.85}
\ding{51} & \ding{51} & \ding{51}  & \textbf{98.37}    & \textbf{8.73}     & \textbf{0.039}   & \textbf{54.92}     & \textbf{0.42}  \\
\bottomrule
\end{tabular}
}
\caption{Ablation studies on the progressive addition of design components under the generation experiment setting.}
\label{tab:ablation on generation 1}
\end{table*}

\begin{figure}[t]
    \centering
    \begin{minipage}[t]{0.47\textwidth}
        \centering
        \footnotesize
        \vspace{0pt}
        \resizebox{\linewidth}{!}{%
        \renewcommand{\arraystretch}{1.2}
        \begin{tabular}{cccc}
        \toprule
        \textbf{Ablation} & \textbf{CLIP}$\uparrow$ & $\textbf{FD}_{\textbf{incep}}\downarrow$ & $\textbf{KD}_{\textbf{incep}}\downarrow$ \\
        \midrule
        w/o DA-SLAT & 97.94  & 10.96 & 0.050    \\
        w/o SMDiT    & 98.10  & 10.03 & 0.047     \\
        w/o MARoPE   & \underline{98.15} &  \underline{9.62}  & \underline{0.043}  \\
        \rowcolor[gray]{0.85}
        Baseline & \textbf{98.37}  & \textbf{8.73} & \textbf{0.039}  \\
        \bottomrule
        \end{tabular}
        }
        \captionof{table}{Ablation study on the impact of each component under the generation experiment setting. \textit{Baseline} indicates the setting where all components are included.}
        \label{tab:ablation on generation 2}
    \end{minipage}
    \hspace{0.01\textwidth}
    \begin{minipage}[t]{0.47\textwidth}
        \centering
         \vspace{0pt}
        \setlength{\abovecaptionskip}{0.6em}
        \includegraphics[width=\linewidth]{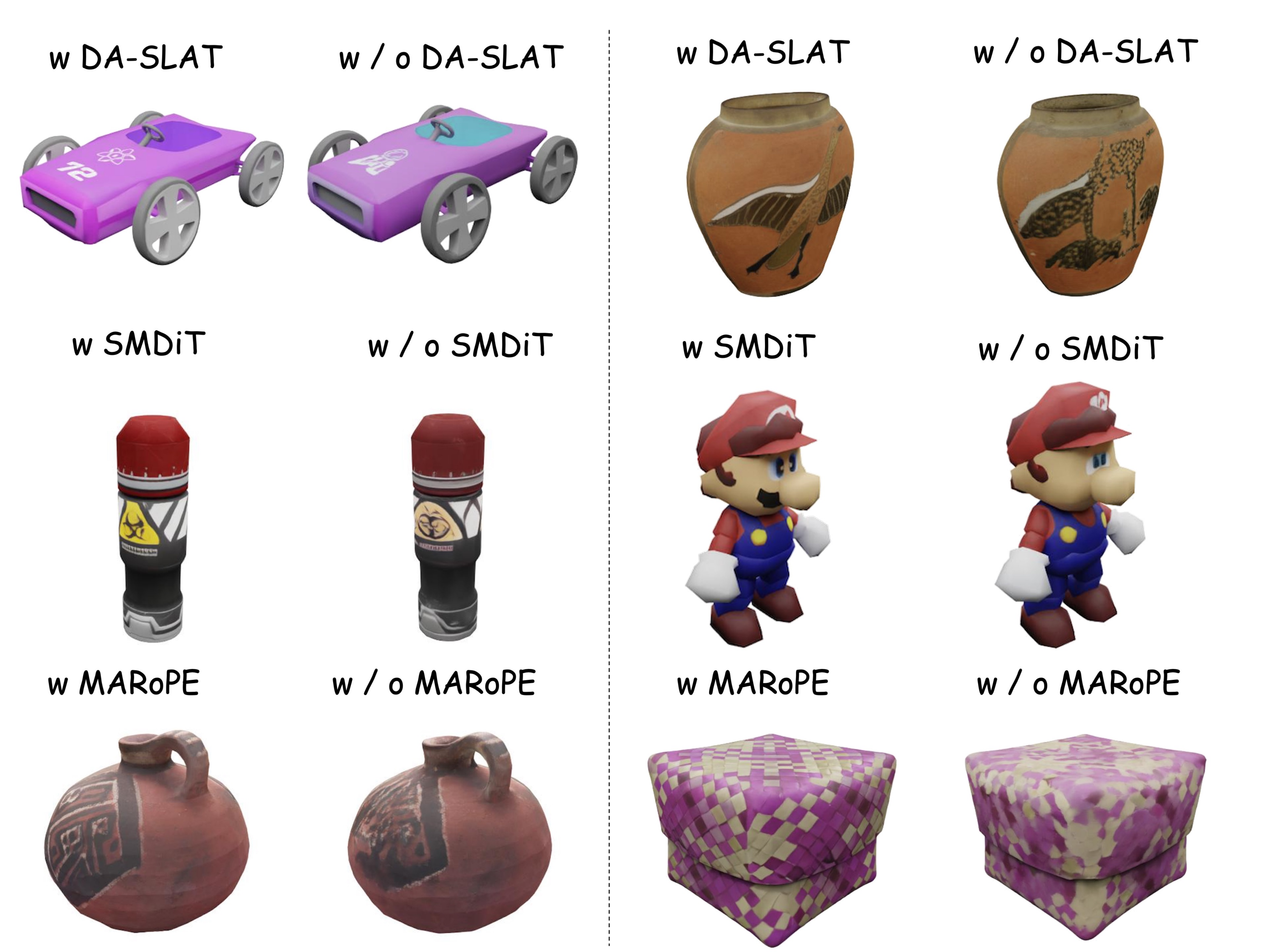}
        \caption{Visualization on the influence of each design component during generation. We disable individual module: the DA-SLAT, the SMDiT architecture, and the MARoPE design, to isolate and understand their respective impact.}
        \label{fig:ablation_gen}
    \end{minipage}
\vspace{-1em}
\end{figure}

\noindent\textbf{Sparse-structure-aware diffusion.}
Finally, we inspect all our design choices under the image-to-3D generation setting to validate their effectiveness. To ensure a comprehensive and robust analysis, we conduct two complementary ablation studies: one that incrementally adds each proposed component to a minimal baseline, and another that removes each component individually from the full configuration to quantify their respective contributions to the final performance.

Quantitative results summarized in ~\cref{tab:ablation on generation 1} yield two key insights: (1) When added to the baseline, DA-SLAT immediately improves all generation metrics, which confirms that its superior reconstruction capability and structured latent space directly benefit the generation quality. (2) SMDiT and MARoPE provide complementary gains: SMDiT mitigates challenges associated with cross-modal alignment, while MARoPE further boosts performance. 
The component-wise removal study in~\cref{tab:ablation on generation 2} and its visualizations in ~\cref{fig:ablation_gen} corroborate these empirical findings. DA-SLAT and SMDiT are more influential and contribute more to appearance details modeling, while MARoPE alleviates smoothing artifacts, enhancing accurate alignment. In summary, the full configuration achieves the best results on all evaluated benchmarks, demonstrating that the synergistic combination of these components is key to our method's performance.

%% file: section/5_conclusion.tex
\section{Conclusion}
\label{sec:conclusion}
\enlargethispage{2\baselineskip}
In this paper, we introduce FLUX3D framework for high-fidelity image-to-3DGS generation using sparse voxel representations. We first show that integrating 2D diffusion features into 3D representation learning significantly boosts reconstruction performance. We further propose SMDiT and MARoPE, tailored for sparse voxel structures, to improve cross-modal alignment between 3D latents and 2D image conditions during diffusion. These components jointly enhance the model’s ability for detailed appearance modeling, producing 3D assets with sharp structural details and intricate textures.

%% file: section/X_suppl.tex



\begin{figure}[t]
    \centering
    \begin{minipage}[t]{0.47\textwidth}
        \centering
        \includegraphics[width=\linewidth]{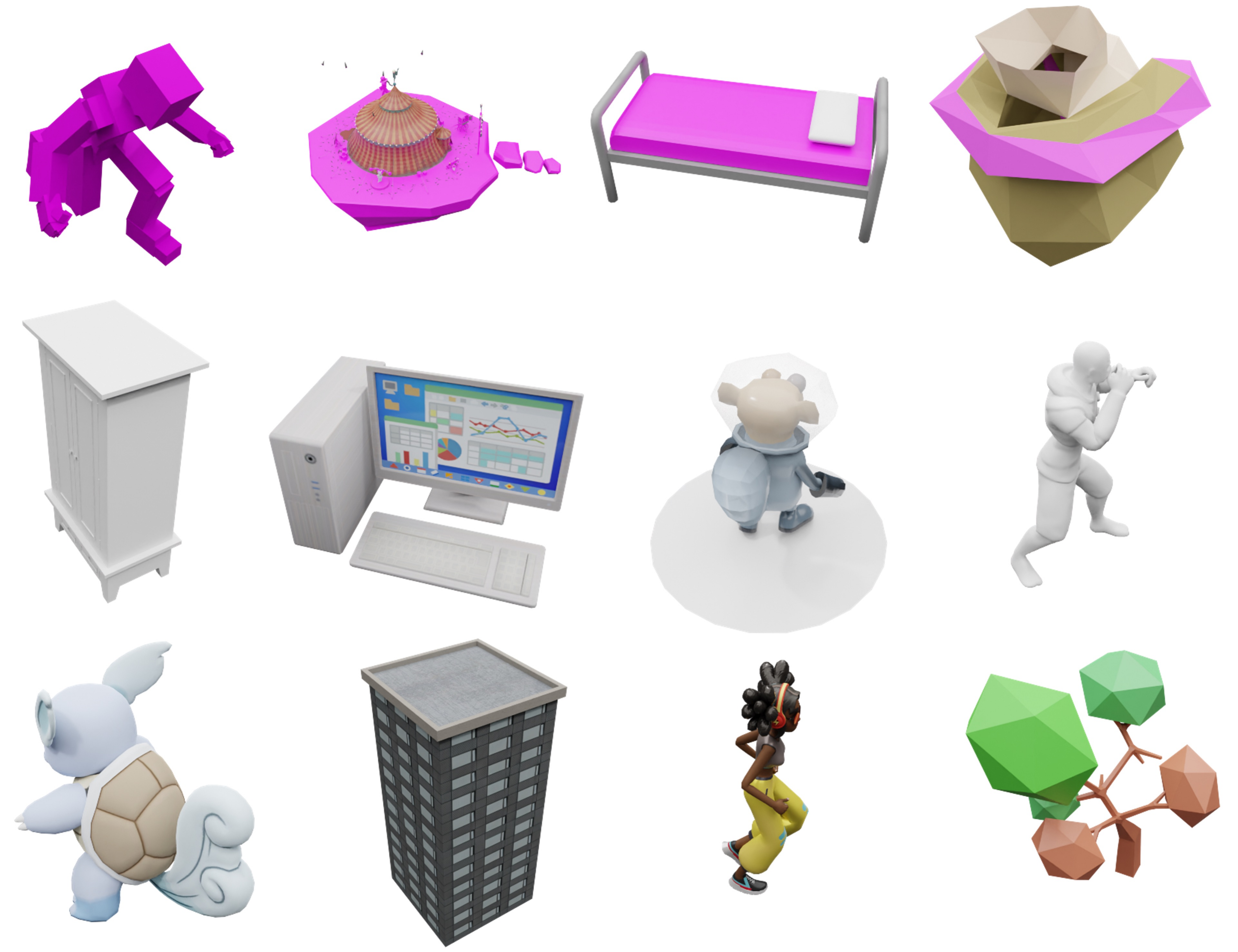}
        \caption{Examples from ObjaverseXL-Github. The first and second rows are assets with missing textures, where the rendering engine will automatically render such areas as pink or white. The third row are assets with normal display.}
        \label{fig:bad data example}
    \end{minipage}
    \hspace{0.01\textwidth}
    \begin{minipage}[t]{0.47\textwidth}
        \centering
        \includegraphics[width=\linewidth]{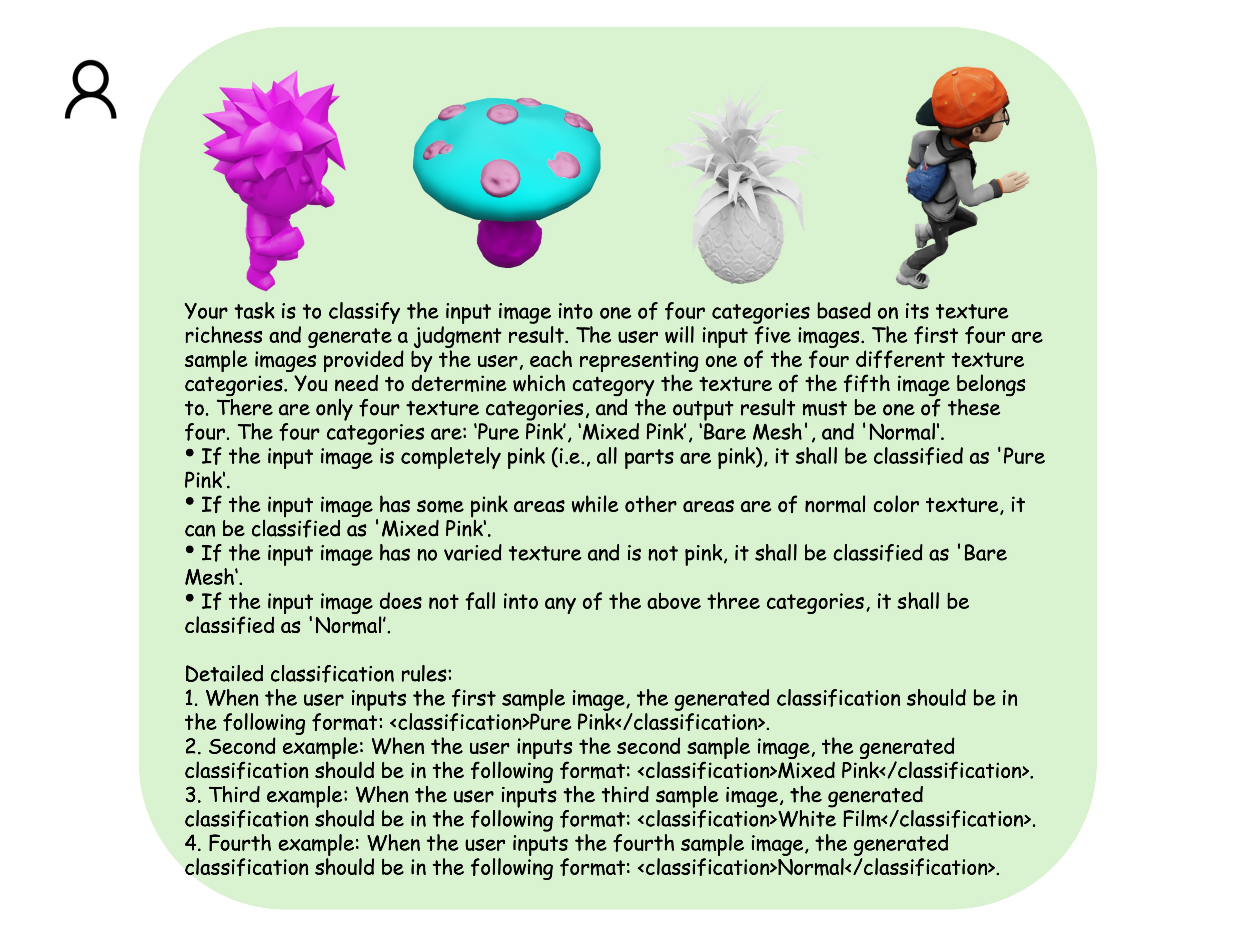}
        \caption{An example of VLM curation process. We classify the 3D assets into four categories: pure pink, mixed pink, bare mesh, and normal assets.}
        \label{fig:vlm curation}
    \end{minipage}
\end{figure}

\section{Additional Implementation Details}

\subsection{Data Preparation Details}
\label{app:data prepare}
To ensure the quality of training data, we conduct a systematic curation process. Firstly, following the practice in~\cite{xiang2025structured}, we render 10 images from uniformly distributed spherical viewpoints around each 3D object. Then, we employ a pretrained aesthetic assessment model to calculate scores across these rendered views and use their average to evaluate the 3D object's visual quality. However, we notice that some 3D assets with missing textures also exhibit relatively high aesthetic scores. As shown in \cref{fig:bad data example}, the rendering engine will automatically render areas with missing textures as pink or white. This phenomenon is particularly evident in the Objaverse-XL~\cite{deitke2023objaverse} dataset, sourced from GitHub. To effectively mitigate the impact of training data bias, we employ an off-the-shelf vision language model to filter out objects in this category. After filtering, about 360K 3D assets remain, which comprise our training dataset. ~\cref{fig:vlm curation} illustrates an example of the curation process, including the designed prompt.

\subsection{Architecture Details}
We show the architecture and configuration of FLUX3D in~\cref{tab:arch config}. For training, we use the AdamW optimizer with a learning rate of $1\times10^{-4}$ and a batch size of 8 per GPU on 8 NVIDIA A100 GPUs, training the SMDiT and decoder for 200K and 100K steps, respectively. The loss weights follow the TRELLIS implementation, with $\lambda_{\text{ssim}}=0.2$, $\lambda_{\text{lpips}}=0.1$, and $\lambda_{\text{vol}}=\lambda_{\alpha}=10^{-3}$. At inference, we use 50 sampling steps. Code and pretrained weights will be released.

\begin{table}[h]
\centering

\begin{tabular}{c|c|c}
\toprule
\textbf{Configuration}  & \textbf{Decoder} & \textbf{SMDiT} \\
\midrule
\# Layers & 12 & 24 \\
\# Heads & 12 & 16 \\
Channel Size & 768 & 1156 \\
\midrule
\# Parameters & 85 M & 820 M \\
\bottomrule
\end{tabular}
\vspace{10pt}
\caption{Configuration of FLUX3D architecture}
\label{tab:arch config}
\end{table}

\section{Additional Quantitative Results}
\subsection{Geometric Reconstruction}
To further verify the influence of input feature selection, we also conduct experiments on geometry reconstruction. Specifically, we implement a mesh decoder based on FlexiCubes~\cite{shen2023flexible}:
\begin{equation}
    \bm{\mathcal{D}_{\text{mesh,flux3d}}}: \{(\bm{f_{i,flux}}, \bm{p_i})\}^L_{i=1} \to \{\{(\bm{w}_i^j, \bm{d}_i^j)\}_{j=1}^{64}\}_{i=1}^L
\end{equation}
where $\bm{w}_i^j \in \mathbb{R}^{16} $ are the flexible parameters and $\bm{d}_i^j \in \mathbb{R}^{8}$ are the signed distance values for the corresponding voxel's eight vertices. After decoding, we extract meshes from 0-level iso-surfaces, and the training objective is computed between rendered depth maps $\bm{D}$, normal maps $\bm{N}$, RGB images $\bm{C}$, and their ground truths:
\begin{equation}
    \begin{aligned}
        \mathcal{L}_{\text{total}} &= \mathcal{L}_{\text{geo}} + \lambda_{\text{color}} \mathcal{L}_{\text{color}} + \lambda_{\text{reg}} \mathcal{L}_{\text{reg}} \\
        \mathcal{L}_{\text{geo}} &= \mathcal{L}_1(\bm{M}) + \mathcal{L}_{\text{Huber}}(\bm{D}) + \mathcal{L}_{\text{recon}}(\bm{N}_m), \\
        \mathcal{L}_{\text{color}} &= \mathcal{L}_{\text{recon}}(\bm{C}) + \mathcal{L}_{\text{recon}}(\bm{N})
    \end{aligned}
\end{equation}
where $\mathcal{L}_{\text{recon}}$ is defined identically to that in 3DGS reconstruction, and $\mathcal{L}_{\text{reg}}$ is the regularization term that ensures plausible mesh extraction.

For geometry quality evaluation, we use Chamfer Distance (CD) to assess shape accuracy, and compute PSNR, and LPIPS on normal maps to evaluate surface details. As shown in ~\cref{tab:geo_recon_exp}, our method outperforms the baselines across all evaluated metrics. However, it's worth noting that the improvement in geometry quality is not as significant as it appears in appearance fidelity, indicating that the input feature volume primarily affects appearance.

\begin{table}[t]
\centering
\footnotesize 

\renewcommand{\arraystretch}{1.2}
\begin{tabular}{cccc}
\toprule
\textbf{Method} & \textbf{CD}$\downarrow$ & \textbf{PSNR-N}$\uparrow$ & \textbf{LPIPS-N}$\downarrow$ \\
\midrule
TRELLIS  &  0.004256  & 30.37  & 0.04216   \\
\rowcolor[gray]{0.85}
Ours (Enc-Dec) &  \underline{0.004234}  & \underline{30.91}  & \underline{0.04072} \\
\rowcolor[gray]{0.85}
Ours (Dec-Only) & \textbf{0.004198} & \textbf{31.13}  & \textbf{0.04021} \\
\bottomrule
\end{tabular}

\vspace{10pt}
\caption{Quantitative evaluation of 3D reconstruction in terms of geometry quality on Toys4k. \textbf{Bold} and \underline{underline} denote the best and the second-best results.}
\label{tab:geo_recon_exp}
\end{table}

\begin{table*}[t]
\centering

\resizebox{\linewidth}{!}{
\renewcommand{\arraystretch}{1.2}
\begin{tabular}{c|ccc|cccccc}
\toprule
\textbf{Method}  & 
\textbf{SSIM}$\uparrow$ & 
\textbf{PSNR}$\uparrow$ & 
\textbf{LPIPS}$\downarrow$ &
\textbf{CLIP}$\uparrow$ &
$\textbf{FD}_{\textbf{incep}}\downarrow$ &
$\textbf{KD}_{\textbf{incep}}(\%)\downarrow$ &
$\textbf{FD}_{\textbf{dinov2}}\downarrow$ &
$\textbf{KD}_{\textbf{dinov2}}(\%)\downarrow$ \\
\midrule
InstantMesh & 0.9232 & 22.87 & 0.07351 &  89.95  &  23.03  &  1.33  & 181.04   &  12.36 \\
3DTopia-XL  & 0.9136 & 20.52 & 0.1006 &  87.31  &  34.49  &  0.56  &  263.01  &  30.62  \\
Shap-E  & 0.8732 & 17.09 & 0.1364 &  85.32  &  63.76  &  4.53  &  400.21  &  66.78  \\
TRELLIS-NeRF  & 0.9521 & 25.62 & 0.04553 &  97.62    &  11.27    & \underline{0.051}   &  62.34  &  0.66  \\
\midrule
\rowcolor[gray]{0.85} 
Ours (Enc-Dec)  &  \underline{0.9615}  &  \underline{26.08}  &   0.03958  & \textbf{98.40}    & \underline{8.81}     & \textbf{0.039}   & \underline{58.43}     & \underline{0.49} \\
\rowcolor[gray]{0.85}
Ours (Dec-Only)  & \textbf{0.9653}  &  \textbf{26.26}  &   \textbf{0.03509}  & \underline{98.37}    & \textbf{8.73}     & \textbf{0.039}   & \textbf{54.92}     & \textbf{0.42} \\
\bottomrule
\end{tabular}
}
\vspace{10pt}
\caption{Quantitative evaluation of image-conditioned 3D generation on Toys4k dataset with non-3DGS-based methods.}
\label{tab:suppl_generation_exp}
\end{table*}

\subsection{Additional Generation Experiments}
To ensure comprehensive evaluation, we not only compare the performance differences among 3DGS-based methods but also conduct cross-type comparisons with mainstream appearance modeling methods that adopt non-3DGS output formats.This includes NeRF-based methods like Shape-E~\cite{jun2023shap}, TRELLIS-NeRF~\cite{xiang2025structured}, and textured mesh, including InstantMesh~\cite{xu2024instantmesh},  3DTopia-XL~\cite{chen20253dtopia}. Same as in Tab.(3), we compute the SSIM, PSNR, and LPIPS between 2D renderings of the generated and ground-truth 3D assets, and also report CLIP Score, Fréchet Distance (FD), and Kernel Distance (KD).

\subsection{Computational Cost and Layout Robustness}
\noindent\textbf{Runtime and memory.} We benchmark the runtime and VRAM consumption of our encoder-decoder and decoder-only variants in the full image-to-3D pipeline on a single A100 GPU (Tab.~\ref{tab:suppl_runtime}). The two variants differ only in the latent-to-hidden input projection (8 vs.~16 channels), which contributes negligible computational overhead during inference.
\begin{table}[t]
\centering
\footnotesize
\setlength{\tabcolsep}{8pt}
\begin{tabular}{lcc}
\toprule
\textbf{FLUX3D Variants} & \textbf{Infer (s)} & \textbf{VRAM (GB)} \\
\midrule
Enc-Dec & 5.10 & 5.18 \\
Dec-Only & 5.11 & 5.18 \\
\bottomrule
\end{tabular}
\vspace{10pt}
\caption{Runtime and VRAM of our two variants in the full image-to-3D pipeline on a single A100 GPU.}
\label{tab:suppl_runtime}
\end{table}

\noindent\textbf{Layout sensitivity.} In Tab.~3 of the main paper, all sparse-voxel-based methods are evaluated as conditional stage-2 generation under the same sparse voxel layout, while other baselines use their original full pipelines. To assess sensitivity to layout quality, we re-run FLUX3D (Dec-Only) and TRELLIS in the full image-to-3D pipeline using the sparse voxel layout predicted by TRELLIS's publicly released stage-1 checkpoint (stage-2 weights unchanged), as reported in Tab.~\ref{tab:suppl_layout}. Although absolute metrics drop under noisy predicted layouts, FLUX3D degrades less and widens its lead over TRELLIS ($\text{FD}_i$ advantage $23\%\!\to\!33\%$, $\text{FD}_d$ $14\%\!\to\!21\%$), suggesting that our gains stem from the FLUX prior and model design rather than from the sparse voxel layout itself.
\begin{table}[t]
\centering
\footnotesize
\setlength{\tabcolsep}{4pt}
\begin{tabular}{lccccc}
\toprule
\textbf{Method} & \textbf{PSNR}$\uparrow$ & \textbf{LPIPS}$\downarrow$ & \textbf{CLIP}$\uparrow$ & $\textbf{FD}_{\textbf{i}}\downarrow$ & $\textbf{FD}_{\textbf{d}}\downarrow$ \\
\midrule
TRELLIS & 20.32 & 0.104 & 94.22 & 18.71 & 129.78 \\
FLUX3D & \textbf{22.32} & \textbf{0.0715} & \textbf{95.57} & \textbf{12.61} & \textbf{102.04} \\
\bottomrule
\end{tabular}
\vspace{10pt}
\caption{Layout sensitivity. Full image-to-3D pipeline using the sparse voxel layout predicted by TRELLIS's stage-1 checkpoint.}
\label{tab:suppl_layout}
\end{table}

\noindent\textbf{FLUX prior vs.~model design.} To disentangle the contribution of the FLUX prior from that of our architectural design, we provide a more detailed ablation that cumulatively adds each factor to the TRELLIS baseline one at a time, as shown in Tab.~\ref{tab:suppl_cumulative}. Both the FLUX prior and our model design (Dec-Only + SMDiT + MARoPE) yield consistent improvements across all metrics, and the model design contributes the larger share: a $16.6\%$ $\text{FD}_i$ gain versus $7.3\%$ from the FLUX prior. This indicates that the improvements stem more from the architectural design than from the FLUX prior. Notably, FLUX feature extraction uses a frozen pretrained encoder comparable to TRELLIS's DINOv2 and thus adds no computational overhead.
\begin{table}[t]
\centering
\footnotesize
\setlength{\tabcolsep}{3pt}
\resizebox{\linewidth}{!}{
\begin{tabular}{lccccc}
\toprule
\textbf{Setting} & \textbf{PSNR}$\uparrow$ & \textbf{LPIPS}$\downarrow$ & \textbf{CLIP}$\uparrow$ & $\textbf{FD}_{\textbf{i}}\downarrow$ & $\textbf{FD}_{\textbf{d}}\downarrow$ \\
\midrule
TRELLIS (DINOv2 + Enc-Dec + DiT) & 25.48 & 0.0439 & 97.92 & 11.29 & 63.66 \\
+ FLUX feat. & 25.74 & 0.0416 & 98.04 & 10.47 & 61.98 \\
+ Dec-Only & 25.84 & 0.0399 & 98.09 & 10.12 & 61.26 \\
+ SMDiT + MARoPE & \textbf{26.26} & \textbf{0.0351} & \textbf{98.37} & \textbf{8.73} & \textbf{54.92} \\
\bottomrule
\end{tabular}
}
\vspace{10pt}
\caption{Cumulative ablation. Each factor is added to the TRELLIS baseline one at a time.}
\label{tab:suppl_cumulative}
\end{table}

\noindent\textbf{Multi-view consistency.} To verify that the improvement is not view-specific, we render 24 views over the full $360^\circ$ azimuth on the Toys4K dataset and report the per-view PSNR against ground truth in Tab.~\ref{tab:suppl_mvc}. FLUX3D achieves both higher accuracy (higher mean PSNR) and better cross-view consistency ($18\%$ lower standard deviation) than TRELLIS.
\begin{table}[t]
\centering
\footnotesize
\setlength{\tabcolsep}{8pt}
\begin{tabular}{lcc}
\toprule
\textbf{Method} & \textbf{Mean PSNR}$\uparrow$ & \textbf{Std}$\downarrow$ \\
\midrule
TRELLIS & 25.08 & 1.76 \\
FLUX3D & \textbf{25.96} & \textbf{1.44} \\
\bottomrule
\end{tabular}
\vspace{10pt}
\caption{Multi-view consistency. Per-view PSNR over 24 views across the full $360^\circ$ azimuth on Toys4K.}
\label{tab:suppl_mvc}
\end{table}

\section{Additional Qualitative Results}
Please refer to our demo video for additional qualitative results and comparisons with SOTA methods.

\section{Limitations and Future works}
While our model demonstrates superior performance on 3DGS
generation, there exist some limitations that remain to be explored. First, the appearance modeling capability for subjects that have semantic meaning, such as texts and logos, still has gap with 2D methods. A potential solution is to collect more comprehensive and high-quality data to improve its capability. Second, the construction of input feature volume relies heavily on multiview renderings, a possible exploration is to see if there are better construction strategies or utilize the information from other modalities like image or video to help improve representation learning quality.